\documentclass[conference,compsoc]{IEEEtran}

\usepackage[utf8]{inputenc} 
\usepackage[T1]{fontenc}
\usepackage{url}
\usepackage{ifthen}
\usepackage{cite}
\usepackage[cmex10]{amsmath} 
                             
\usepackage{amsmath,amssymb, graphicx}
\usepackage[english]{babel}
\usepackage[ruled,linesnumbered]{algorithm2e}
\usepackage{algorithmic}
\usepackage{paralist} 
\usepackage{color}
\usepackage{multirow}
\usepackage{xspace}
\usepackage{tabularx}
\usepackage{rotating}
\usepackage{hyperref}
\usepackage[mathscr]{eucal} 
\usepackage{amsbsy} 
\newcommand{\tensor}[1]{\boldsymbol{\mathscr{#1}}}   
\usepackage{subfigure}
\usepackage{subcaption}
\usepackage{booktabs}
\usepackage{xcolor}
\usepackage{tikz}
\usetikzlibrary{snakes}
\usepackage{multirow}
\usepackage{epstopdf}
\usepackage{balance}
\usepackage{tablefootnote}
\usepackage{empheq} 
\usepackage{moresize}
\usepackage{enumitem}
\usepackage{graphicx}
\usepackage[utf8]{inputenc}
\setlist{nolistsep}
\usepackage{layouts}
\usepackage{float}
\usepackage[font=small,labelfont=bf,skip=0pt]{caption}

\DeclareCaptionType{copyrightbox}
\usepackage{url}

\newcounter{ALC@tempcntr}
\newcommand{\hide}[1]{}

\newcommand{\ben}{\begin{enumerate*}}
\newcommand{\een}{\end{enumerate*}}
\newcommand{\bit}{\begin{itemize*}}
\newcommand{\eit}{\end{itemize*}}

\newcommand{\T}[1]{\boldsymbol{\mathscr{#1}}}
\newcommand{\mat}[1]{\mathbf{#1}}
\newcommand{\vect}[1]{\mathbf{#1}}

\def\x{{\mathbf x}}

\def\S{{\mathbf{S}}}

\begin{document}

 \author{\IEEEauthorblockN{Shaan Pakala}
 \IEEEauthorblockA{Dept. of CSE\\
 UC Riverside\\
 shaan.pakala@email.ucr.edu}
 \and
 \IEEEauthorblockN{Bryce Graw}
 \IEEEauthorblockA{Dept. of CIS\\
 San Diego Mesa Community College\\
 bgraw@student.sdccd.edu}
 \and
  \IEEEauthorblockN{Dawon Ahn}
 \IEEEauthorblockA{Dept. of CSE\\
 UC Riverside\\
 dahn017@ucr.edu}
 \and
   \IEEEauthorblockN{Tam Dinh}
 \IEEEauthorblockA{Dept. of CS\\
 Cal Poly Pomona\\
 tamdinh@cpp.edu}
 \and
 \IEEEauthorblockN{Mehnaz Tabassum Mahin }
 \IEEEauthorblockA{Dept. of CSE\\
 UC Riverside\\
mehnaztabassum.mahin@email.ucr.edu }
\and
 \IEEEauthorblockN{Vassilis Tsotras}
 \IEEEauthorblockA{Dept. of CSE\\
UC Riverside\\
 tsotras@cs.ucr.edu}
 \and
 \IEEEauthorblockN{Jia Chen}
 \IEEEauthorblockA{Dept. of ECE\\
 UC Riverside\\
 jiac@ucr.edu}
 \and
 \IEEEauthorblockN{Evangelos E. Papalexakis}
 \IEEEauthorblockA{Dept. of CSE\\
 UC Riverside\\
 epapalex@cs.ucr.edu}
 }

\title{Automating Data Science Pipelines with Tensor Completion}

\maketitle
\begin{abstract} 

Hyperparameter optimization is an essential component in many data science pipelines and typically entails exhaustive time and resource-consuming computations in order to explore the combinatorial search space. Similar to this problem, other key operations in data science pipelines exhibit the exact same properties. Important examples are: neural architecture search, where the goal is to identify the best design choices for a neural network, and query cardinality estimation, where given different predicate values for a SQL query the goal is to estimate the size of the output. 
In this paper, we abstract away those essential components of data science pipelines and we model them as instances of tensor completion, where each variable of the search space corresponds to one mode of the tensor, and the goal is to identify all missing entries of the tensor, corresponding to all combinations of variable values, starting from a very small sample of observed entries.
In order to do so, we first conduct a thorough experimental evaluation of existing state-of-the-art tensor completion techniques and introduce domain-inspired adaptations (such as smoothness across the discretized variable space) and an ensemble technique which is able to achieve state-of-the-art performance.
We extensively evaluate existing and proposed methods in a number of datasets generated corresponding to (a) hyperparameter optimization for non-neural network models, (b) neural architecture search, and (c) variants of query cardinality estimation, demonstrating the effectiveness of tensor completion as a tool for automating data science pipelines. Furthermore, we release our generated datasets and code in order to provide benchmarks for future work on this topic.

\end{abstract}

\section{Introduction}
\label{sec:intro}

In many different data science pipelines, it is necessary to automate the pipeline's design such as choosing the optimal hyperparameters for a machine learning model, searching for optimal architecture for a deep neural network, and predicting the (distinct) output cardinality of all predicate values for SQL  queries.
Unfortunately, this usually requires exhaustive computation of all candidate combinations of designs resulting in expensive computing time and resource cost, exponential with respect to the number of search components, e.g., hyperparameters and SQL query predicates.

Finding the optimal hyperparameter configuration of a machine learning model is commonly known as hyperparameter optimization. Standard approaches include grid search and random search \cite{bergstra2011algorithms}. Grid search (such as GridSearchCV \cite{scikit-learn}) exhaustively trains and evaluates a machine learning model's performance using a grid of hyperparameter combinations. The grid represents the hyperparameters on each axis, and the different axis values are the predefined values that hyperparameter takes. These predefined hyperparameter combinations are all trained and evaluated on a downstream task \cite{bellman1959adaptive}. As the number of hyperparameters grow, the number of combinations of values they take grow exponentially, making an exhaustive search very difficult. An exhaustive search becomes impractical with even more intricate deep learning model architectures as well.
Another simple approach, random search, evaluates a model on the hyperparameters that are drawn independently from a predefined distribution, such as uniform distribution. 
However, these algorithms often do not converge to the optimal configuration \cite{bergstra2011algorithms}.

In the domain of hyperparamter optimization, there have been efforts to address the computational challenges of the standard methods posed by manual programming and testing. Bayesian optimization \cite{wu2019hyperparameter} uses probabilistic models such as random forests, Gaussian processes, and gradient boosting to decide which data sample to be fed for the evaluation in each iteration \cite{van2021hyperboost,hutter2011sequential,olson2018data}. Stochastic population-based optimization methods, evolution strategies, iteratively find hyperparameter configurations with high fitness values, i.e., the (inverted) generalization error \cite{he2021automl,beyer2002evolution,coello2007evolutionary}. Other advanced frameworks include Hyperband \cite{li2018hyperband}, iterated racing \cite{birattari2010f}, and Gradient-based optimization methods \cite{baydin2018automatic,lorraine2020optimizing}. The aforementioned methods have been shown promising results in searching for good hyperparameter configurations, while none of them are designed to explicitly take advantage of the underlying low-rank structure as, in practice, there is often a small subset of the whole set of hyperparameters that influence the downstream task performance, and several values of the same hyperparameter, e.g., learning rate,  most likely yield to similar or the same performance \cite{bergstra2012random,bischl2023hyperparameter}. 
There exists work on approaching hyperprameter optimization problem as a tensor completion task which assumes the tensor being low rank \cite{deng2022new, yang2020automl, rebelloefficient}.

\begin{figure*}[!ht]
    \begin{center}
        \includegraphics[width = \textwidth]{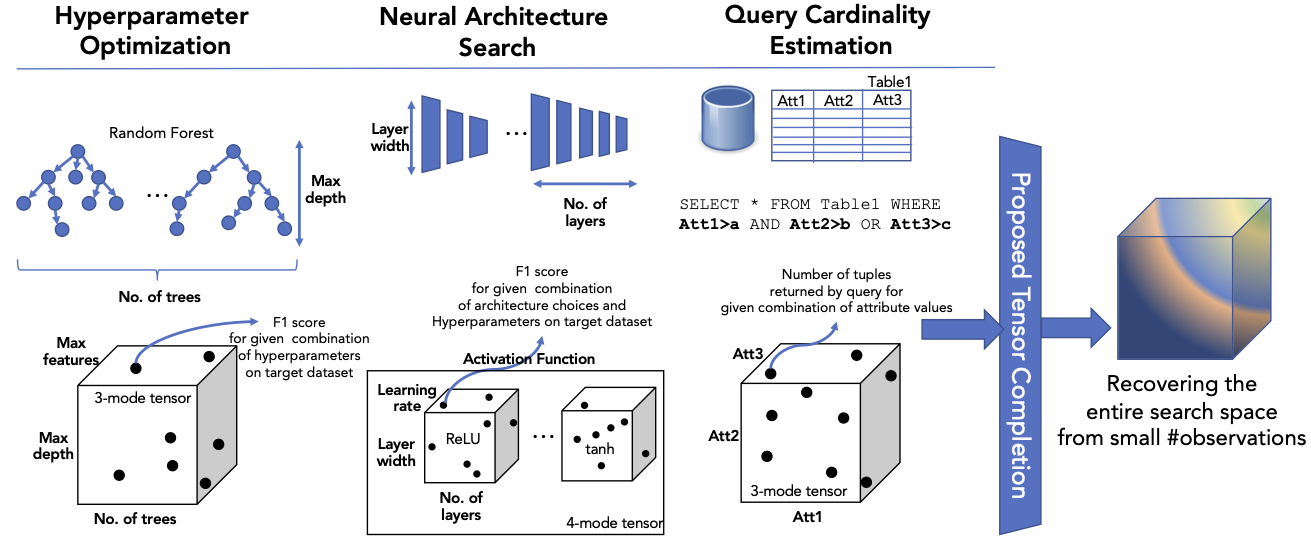}
        \caption{In this work we unify a number of combinatorial and highly computationally intense data science tasks, such as hyperparameter optimization, neural architecture search, and query cardinality estimation, under the umbrella of tensor completion. We conduct a thorough and extensive study of existing tensor completion methods and propose a novel method for accurately recovering the entire search space in those data science tasks from a small number of observations, towards automating data science pipelines.}
    \end{center}
\end{figure*}

However, the existing methods are limited to either suggesting a single optimal hyperparameter configuration, depending on strong underlying statistical assumptions, or being constrained to low-rank structure. In the area of data science pipeline design, one may be interested in multiple optimal hyperparameter configurations or knowing the outcome of every configuration, or can not make any statistical or structural assumptions. In database, users tend to care about the cardinality of every query. To this end, we offer a generic model for automating data science pipelines through tensor completion. We model the outputs of the data science pipeline design (e.g., cardinality of SQL query searches and machine learning model evaluation scores) as the entries of a tensor, and each search variable (e.g., hyperparameter and a query predicate value) as one mode of the tensor. Given a very small fraction of observed entries, our goal is to estimate all missing entries of the tensor.

In this paper, we thoroughly explore a broad spectrum of the existing state-of-the-art (SOTA) tensor completion techniques including  CPD, TuckER\cite{balazevic2019tucker}, CoSTCo \cite{liu2019costco}, and NeAT\cite{ahnneural}, and apply them in automating data science pipelines. Leveraging the advances of CPD, we also propose a new tensor completion method, namely CPD-S, by enforcing the smoothness constraints on the latent CPD components since any two similar values of the same search variable most likely lead to similar design output. Furthermore, to fully take advantage of the individual SOTA models, we propose ensemble methods. Specifically, we consider: 1) straight-forward schemes such as taking the mean and median of several prediction results corresponding to the same tensor entry  from the same model (CPD-S or CoSTCo) with different pre-defined tensor ranks, and 2) more advanced framework, i.e., learning the contributions of the results from multiple CPD-S or CoSTCo through a neural network such as multi-layer preceptron and aggregating the results nonlinearly to form a single prediction result for every design configuration.

In this work, we make several distinct contributions in sparse tensor completion for surrogate modeling as follows:
\begin{itemize}
    \item {\bf Broad Data Science Applications}: we take a broader view and demonstrate how this can generalize to a variety of data science problems in machine learning and database management.
    \item {\bf Investigating Tensor Completion Variants}: we explore the performance of different tensor completion variants and identify pros and cons.
    \item {\bf Leveraging and Exploring Tensor Modeling}: we explore the expressive power of tensor modeling in exploring ways to further improve surrogate modeling.
    \item {\bf Proposed Methods}: 
    Motivated by the problem structure, we propose applying a smoothness constraint to factor matrices in CPD tensor completion to improve performance for this application.
    Furthermore, we propose an ensemble tensor completion to aggregate the results of several tensor completion methods, for improved performance and reliability.
    \item {\bf Public Code and Benchmark Datasets}: In order to promote reproducibility and follow-up research, we make our implementation and the benchmark datasets  created for (i) hyperparameter optimization, (ii) neural architecture search, and (iii) query cardinality estimation publicly available at \url{https://github.com/shaanpakala/STC_AutoML}.
\end{itemize}

\section{Preliminaries and Problem Formulation}
\label{sec:prelim}

\subsection{Preliminaries}

\subsubsection{Data Science Pipelines}

Below are the three bottleneck tasks within a data science pipeline that we focus on.

\noindent{\bf Hyperparameter Optimization}:
\noindent{We are finding the best combination of hyperparameters \cite{bergstra2011algorithms} for a non-deep learning machine learning model (e.g. K-Nearest Neighbors, Decision Trees). This involves adjusting the values of various hyperparameters of the machine learning model, then training the model and evaluating its performance.}

\noindent{\bf Neural Architecture Search}:
\noindent{This is a similar task to Hyperparameter Optimization, except with Neural Networks (e.g. Dense Neural Networks, Convolutional Neural Networks) \cite{zoph2016neural, jin2019auto, mellor2021neural}. Now there are different hyperparameters, such as layer size or number of layers. The goal here is to find the Neural Network architecture that gives the best performance, by training and evaluating different architectures.}

\noindent{\bf Query Cardinality Estimation}:
\noindent{We want to estimate the cardinality of the output for complex database queries \cite{woltmann2019cardinality, malik2007black}. In this application, we use tensor completion to extract the complex relationships between the predicates of a query, in order to infer the entire output cardinality. In this case, the hyperparameters would be the attribute in the predicates, and the hyperparameter values would be the range for that predicate's attribute.}

\noindent{\bf Query Distinct Cardinality Estimation}:
\noindent{We want to estimate the cardinality of the distinct values of a given attribute in the output for database queries. Here the focus is to infer the cardinality of the distinct values of the output, rather than just the entire output. This is almost identical to Query Cardinality estimation, except the entries in the tensor represent cardinality of distinct values (of a specificied attribute).}

\subsubsection{Surrogate Modeling}

Surrogate modeling in machine learning is used to help guide the optimal hyperparameter search without exhaustively train and evaluating all combinations  \cite{yang2020automl}. In our case, this will be done by estimating the performance (according to an evaluation metric, F1 Score) for all configurations of machine learning models. This will also be applied to infer the cardinalities of various database queries, with their own design configurations.

\subsubsection{Tensors}

Tensors are multidimensional arrays. In other words, a vector is a 1-dimensional tensor, and matrix is a 2-dimensional tensor. We will be looking at tensors of 3 or more dimensions \cite{kolda2009tensor,sidiropoulos2016tensor}. The tensors for our application will also be Sparse Tensors, which are tensors with many missing values. We use boldface italicized letters (e.g. $\tensor{X}$) to denote dense tensors (tensors with no missing values). For sparse tensors, we will use the same notation, with a subscript "$S$". For example, $\tensor{X}_S$ would be a sparse tensor corresponding to dense tensor $\tensor{X}$.

\subsubsection{Tensor Decomposition}

Tensor decomposition is the process of expressing a tensor using smaller factors. For example, a common method of tensor decomposition is the Canonical Polyadic Decomposition (CPD) \cite{kolda2009tensor,sidiropoulos2016tensor}. CPD expresses a tensor as a sum of rank-one tensors. A third-order tensor $\tensor{X} \in \mathbb{R}^{I \x J \x K} $ would be expressed as:
$
\tensor{X} \approx \sum_{r=1}^{R} (\mathbf{a}_r \circ \mathbf{b}_r \circ \mathbf{c}_r),
$
where $\circ$ denotes outer product, $\mathbf{a}_r \in \mathbb{R}^I, \mathbf{b}_r \in \mathbb{R}^J, \text{ and } \mathbf{c}_r \in \mathbb{R}^K$. We represent a decomposition of tensor $\tensor{X}$ as $\tensor{X}_D$.

\subsubsection{Tensor Completion}

Tensor Completion is the process of filling in the missing values of a sparse tensor. There are several forms of Tensor Completion methods, such as:

\noindent{\textbf{Classical Tensor Methods}: CPD \cite{PARAFAC} \& TuckER\cite{balazevic2019tucker}},

\noindent{\textbf{Tensor Network Methods}: Tensor Train \cite{oseledets2011tensor}}, and

\noindent{\textbf{Neural Tensor Methods}: CoSTCo\cite{liu2019costco} \& NeAT\cite{ahnneural}}.

Classical Tensor Methods heavily rely upon the tensor decomposition in order to infer back the missing values. CPD, for example, uses a series of matrix multiplication and addition in order to generate the dense tensor again. 

Tensor Networks \cite{sengupta2022tensor} have been witnessing a steady rise in popularity, as they provide a flexible and modular framework for expressing otherwise complex tensor models using elementary tensor operations such as multilinear maps (which, in a nutshell, describe how a tensor mode is projected from its original space to a new space) in network form. One of the most prominent Tensor Network models is the so-called Tensor Train \cite{oseledets2011tensor}, which we use here as a representative of Tensor Network methods.

Neural Tensor Methods use an additional neural network to infer back the missing values of the tensor. CoSTCo \cite{liu2019costco}, for example, uses a Convolutional Neural Network (CNN) to extract information from very sparse tensors to accurately infer back the missing values.

\subsubsection{Tensor Completion Training}

Training our tensor completion algorithms begin with randomly initializing a decomposition of the dense tensor $\tensor{X}$. Using this random decomposition $\tensor{X}_D$, we can generate an estimation of the dense tensor $\tensor{X}$ according to the corresponding method's algorithm. For example, CPD  uses  matrix products and sums to reconstruct the full tensor, while CoSTCo \cite{liu2019costco}  uses CNNs. After building the estimation of the dense tensor $\tensor{X}$, we  calculate the loss using the Mean Squared Error (MSE):
\begin{equation}
Loss = \text{Mean}((\tensor{X}_S - R(\tensor{X}_D))^2 \cdot M_{\tensor{X}})
\end{equation}

\noindent where $R(\tensor{X})$ represents the full reconstruction of tensor $\tensor{X}$ using only the randomly initialized factors $\tensor{X}_D$.
$M_{\tensor{X}}$ is a boolean/mask tensor, whose values are 1 if that entry in $\tensor{X}_S$ is observed, and 0 if it is unobserved. This is because we only keep the reconstruction for the observed indices of the sparse tensor to calculate our loss. We multiply the loss by the mask, to convert the loss for the missing entries to 0, and the non-missing entries remain the same.

Now we can optimize the randomly initialized decomposition $\tensor{X}_D$ in terms of MSE using Adam \cite{kingma2014adam} with backpropagation. We are essentially iteratively getting closer to a full reconstruction of $\tensor{X}$ such that the entries corresponding to the sparse tensor values are similar to their observed values.

\subsection{Problem Definition}

The general problem we are solving is to be able to efficiently and accurately approximate data science pipelines' outputs across a large combinatorial space of configurations.

\subsubsection{Tensor Completion for Hyperparameter Tuning}

Hyperparameter tuning can be modeled as a tensor, where each axis of the tensor represents a certain hyperparameter we are optimizing. Each of the values of that axis will represent the different values the hyperparameter can take. Each cell will now represent a specific combination of hyperparameter values, which will be filled in using some evaluation metric of the Machine Learning model using that hyperparameter combination. For our purposes, we will be using F1-Score for classification tasks.

This figure demonstrates how we will use only a portion of selected hyperparameter combinations, to infer the entire space of combinations of hyperparameters, using tensor completion. Above is an example of TuckER tensor completion, using Tucker tensor decomposition \cite{balazevic2019tucker} Each of $H_1$, $H_2$, and $H_3$ represents a different hyperparameter that we are tuning, corresponding to that axis of the tensor.

In the figure, the tensor represents hyperparameter combinations for k-Nearest Neighbors (k-NN). Here, one axis represents k, with different values k could take, and another axis represents $p$ (in L$p$ norm distance), with different values p could take. The cells represent a combination of hyperparameters that k-NN could take, using an evaluation metric. We will be using F1 Score for the evaluation metric.

\subsubsection{Tensor Completion for Neural Architecture Search}

Neural Architecture Search will be modeled similarly, where each axis will now represent a different kind of hyperparameter. Some examples are number of layers, layer size, and different activation functions. Each cell will also represent a combination of hyperparameter values, where the value of that cell will also represent some evaluation metric of the Deep Learning model using that hyperparameter combination. We will also be using F1-Score here.

\subsubsection{Tensor Completion for Query Cardinality Estimation}

Query Cardinality Estimation can be modeled as a tensor where each axis represents a certain predicate of a query. Each value of that axis represents a different value that the predicate could take. Now each cell will represent the cardinality of the output that is produced using this combination of predicate values.

\subsubsection{Tensor Completion for Query Distinct Cardinality Estimation}

Very similar to Query Cardinality Estimation, except now we are estimating the cardinality of the distinct values of an attribute in the output of a database query.

\section{Proposed Method}
\label{sec:proposed}


In this work, we investigate the behavior of several types of Classical, Tensor Network, and Neural Tensor Completion models for a variety of tasks.
We also propose applying a smoothness constraint \cite{ahn2021time} on all modes for CPD tensor completion. In addition to these individual tensor completion methods, we propose an ensemble tensor completion model, to aggregate the results of several individual ones.

\subsection{Dataset Generation}
Dataset generation consists of exhaustively computing the outcomes of all combinations (of a prespecified range) of hyperparameters, neural network layers, and queries. From here, we can remove values from this complete tensor to produce our sparse tensor, simulating computing only a fraction of the combinations. Then we can observe how well we are able to infer them again. This will allow us to evaluate the performance of our tensor completion methods for these applications.

\begin{table*}[t]
\ssmall 
\centering
\setlength{\tabcolsep}{2.5pt}
\caption{Hyperparameters \& Ranges \label{tab:hyperparameters} }
\begin{tabular}{lllllll}
\toprule
Tensor File & Tensor Size & H1 & H2 & H3 & H4 & H5 \\
\midrule
Non-Deep Learning & & & & & & \\
\midrule
KNN$\_$car$\_$evaluation$\_$828 & 3x7x2x9x11 & scaler[None,minmax,standard] & PCA[1,2,3,4,5,6,None] & weights['distance','uniform'] & p[1-10] & n$\_$neighbors[1-75] \\
DT$\_$Dermatology$\_$828 & 8x8x8x9 & max$\_$depth[1-15,None] & max$\_$features[1-15,None] & min$\_$samples$\_$leaf[1-100] & - & - \\
RF$\_$Dermatology$\_$828 & 8x8x8x9 & min$\_$samples$\_$leaf[1-75] & max$\_$features[1-15,None] & max$\_$depth[1-15,None] & n$\_$estimators[1-100] & - \\
SVM$\_$Biodeg$\_$905 & 3x2x7x9x11 & scaler[None,minmax,standard] & SMOTE[False,True] & C[0.125-8] & degree[1-15] & max$\_$iter[2-500] \\
SVM$\_$Dermatology$\_$905 & 3x2x7x9x11 & scaler[None,minmax,standard] & SMOTE[False,True] & C[0.125-8] & degree[1-15] & max$\_$iter[2-500] \\
SVM$\_$Alzheimers$\_$905 & 3x2x7x9x11 & scaler[None,minmax,standard] & SMOTE[False,True] & C[0.125-8] & degree[1-15] & max$\_$iter[2-500] \\
DT$\_$Spambase$\_$829 & 3x4x8x7x9 & scaler[None,minmax,standard] & min$\_$impurity$\_$decrease[0-0.1] & min$\_$samples$\_$leaf[1-75] & max$\_$features[1-10,None] & max$\_$depth[1-15,None] \\
RF$\_$Spambase$\_$829 & 3x4x8x7x9 & scaler[None,minmax,standard] & min$\_$impurity$\_$decrease[0-0.1] & min$\_$samples$\_$leaf[1-75] & max$\_$features[1-10,None] & max$\_$depth[1-15,None] \\
ET$\_$Spambase$\_$829 & 3x4x8x7x9 & scaler[None,minmax,standard] & min$\_$impurity$\_$decrease[0-0.1] & min$\_$samples$\_$leaf[1-75] & max$\_$features[1-10,None] & max$\_$depth[1-15,None] \\
GB$\_$Spambase$\_$829 & 3x4x8x7x9 & scaler[None,minmax,standard] & min$\_$impurity$\_$decrease[0-0.1] & min$\_$samples$\_$leaf[1-75] & max$\_$features[1-10,None] & max$\_$depth[1-15,None] \\
\midrule
Neural Architecture Search & & & & & & \\
\midrule
FCNN$\_$Dermatology$\_$829 & 3x6x3x6x5 & scaler[None,minmax,standard] & num$\_$epochs[3-25] & batch$\_$size[16-256] & num$\_$layers[1-10] & hidden$\_$size[32-5000] \\
FCNN$\_$Alzheimers$\_$902 & 3x6x3x6x4 & scaler[None,minmax,standard] & num$\_$epochs[2-25] & batch$\_$size[128-2048] & num$\_$layers[1-15] & hidden$\_$size[32-512] \\
FCNN$\_$car$\_$evaluation$\_$903 & 3x6x3x6x4 & scaler[None,minmax,standard] & num$\_$epochs[2-25] & batch$\_$size[128-2048] & num$\_$layers[1-15] & hidden$\_$size[32-512] \\
FCNN$\_$Dermatology$\_$903 & 3x6x3x6x4 & scaler[None,minmax,standard] & num$\_$epochs[2-25] & batch$\_$size[128-2048] & num$\_$layers[1-15] & hidden$\_$size[32-512] \\
FCNN$\_$Particle$\_$ID$\_$903$\_$02 & 3x6x3x6x4 & scaler[None,minmax,standard] & num$\_$epochs[2-25] & batch$\_$size[128-2048] & num$\_$layers[1-15] & hidden$\_$size[32-512] \\
FCNN$\_$Spambase$\_$902 & 3x6x3x6x4 & scaler[None,minmax,standard] & num$\_$epochs[2-25] & batch$\_$size[128-2048] & num$\_$layers[1-15] & hidden$\_$size[32-512] \\
FCNN$\_$Spambase$\_$905$\_$50 & 3x6x5x5x5 & activation[relu,sigmoid,tanh] & lr[0.0005-0.1] & num$\_$epochs[5-50] & hidden$\_$size[10-250] & num$\_$layers[1-10] \\
FCNN$\_$Biodeg$\_$905 & 3x6x5x5x5 & activation[relu,sigmoid,tanh] & lr[0.0005-0.1] & num$\_$epochs[5-50] & hidden$\_$size[10-250] & num$\_$layers[1-10] \\
FCNN$\_$Particle$\_$ID$\_$903$\_$02 & 3x6x3x6x4 & scaler[None,minmax,standard] & num$\_$epochs[2-25] & batch$\_$size[128-2048] & num$\_$layers[1-15] & hidden$\_$size[32-512] \\
FCNN$\_$Spambase$\_$902 & 3x6x3x6x4 & scaler[None,minmax,standard] & num$\_$epochs[2-25] & batch$\_$size[128-2048] & num$\_$layers[1-15] & hidden$\_$size[32-512] \\
\midrule
Query Cardinality & & & & & & \\
\midrule
AND$\_$AND$\_$801 & 10x10x10 & surname$\_$pcode['B','V','Q','G','L'] & name$\_$pcode$\_$nf['G','I','M','O','K'] & person$\_$id[1e5-1e6] & - & - \\
AND$\_$OR$\_$801 & 10x10x10 & surname$\_$pcode['B','V','Q','G','L'] & name$\_$pcode$\_$nf['G','I','M','O','K'] & person$\_$id[1e5-1e6] & - & - \\
OR$\_$AND$\_$801 & 10x10x10 & surname$\_$pcode['B','V','Q','G','L'] & name$\_$pcode$\_$nf['G','I','M','O','K'] & person$\_$id[1e5-1e6] & - & - \\
OR$\_$OR$\_$801 & 10x10x10 & surname$\_$pcode['B','V','Q','G','L'] & name$\_$pcode$\_$nf['G','I','M','O','K'] & person$\_$id[1e5-1e6] & - & - \\
\midrule
Query Distinct Cardinality & & & & & & \\
\midrule
AND$\_$AND$\_$distinct$\_$817 & 10x10x10 & surname$\_$pcode['B','V','Q','G','L'] & name$\_$pcode$\_$nf['G','I','M','O','K'] & person$\_$id[1e5-1e6] & - & - \\
AND$\_$OR$\_$distinct$\_$817 & 10x10x10 & surname$\_$pcode['B','V','Q','G','L'] & name$\_$pcode$\_$nf['G','I','M','O','K'] & person$\_$id[1e5-1e6] & - & - \\
OR$\_$AND$\_$distinct$\_$817 & 10x10x10 & surname$\_$pcode['B','V','Q','G','L'] & name$\_$pcode$\_$nf['G','I','M','O','K'] & person$\_$id[1e5-1e6] & - & - \\
OR$\_$OR$\_$distinct$\_$817 & 10x10x10 & surname$\_$pcode['B','V','Q','G','L'] & name$\_$pcode$\_$nf['G','I','M','O','K'] & person$\_$id[1e5-1e6] & - & - \\
\bottomrule
\end{tabular}

\vspace{0.1cm}
\normalsize
This table displays each tensor we generated and the hyperparameters or components used, along with its range of values.

The Query Tensors (Query Cardinality \& Distinct Cardinality) are generated on the IMDB dataset \cite{leis2015good}, and each cell represents the attribute along with the values the attribute will be compared against. For example, surname$\_$pcode ['B','V','Q','G','L'] means the predicates are surname$\_$pcode >= 'B', surname$\_$pcode <= 'B', surname$\_$pcode >= 'V', etc.

\end{table*}

\noindent{\bf Non-Deep Learning}
We use scikit-learn \cite{scikit-learn} to exhaustively train and evaluate non-deep learning model hyperparameter combinations. 

\noindent{\bf Neural Architecture Search} We use PyTorch \cite{paszke2019pytorch} to create dense neural networks, to also exhaustively run combinations of hyperparameters for our neural architecture search tensor. 

\noindent{\bf Query Cardinality Estimation} We run queries with combinations of predicate values and generate the cardinalities of the outputs for our Query Cardinality tensors and cardinalities of distinct output values for a given attribute.

A full detailed list of hyperparameters \& components used for each tensor we generated can be found in Table \ref{tab:hyperparameters}.

\subsection{CPD-S: Smooth CP Decomposition}
\begin{figure}[!ht]
	\centering
    \subfigure[SVM Tensor]{
        \includegraphics[width=0.2\textwidth]{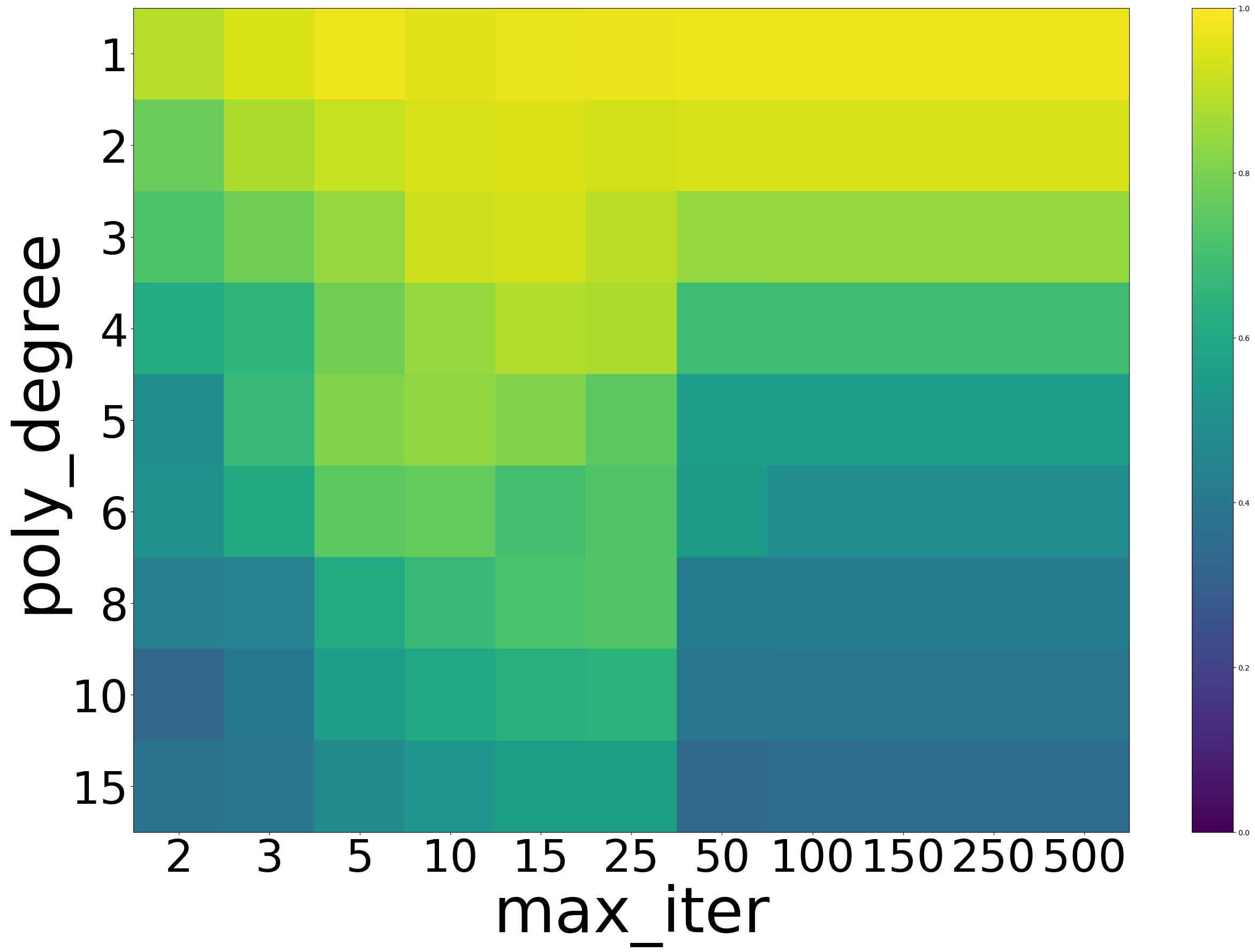}
    }
    \subfigure[Neural Network Tensor]{
        \includegraphics[width=0.2\textwidth]{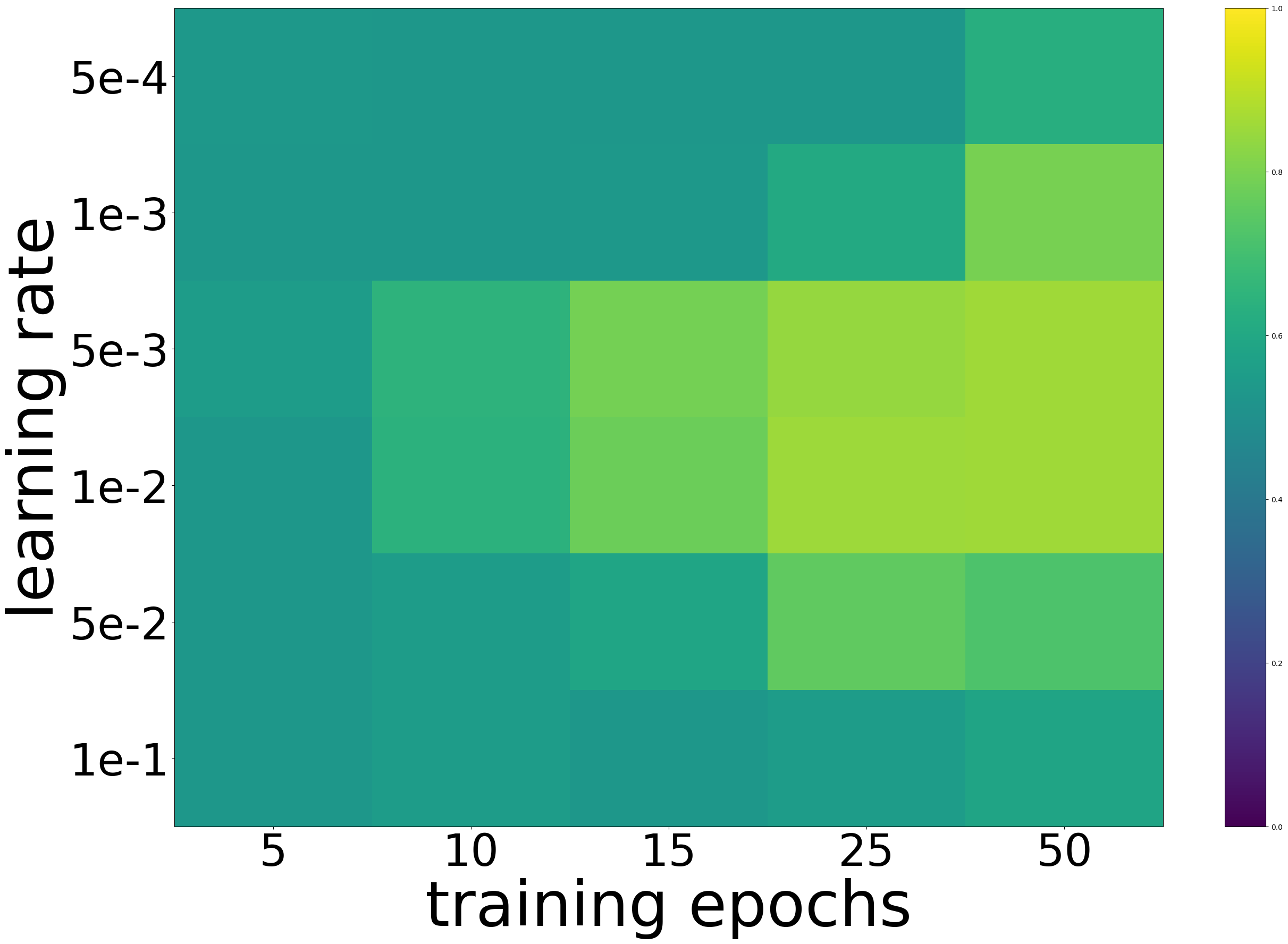} 
    } 
    \caption{Indicative tensor slices where the nature of the hyperparameters involved results in smoothness across those dimensions motivating our proposed smoothness constrained CPD-S method.\label{hp_smoothness}
    }
\end{figure}

The indices of the hyper-parameter tensors are sequentially ordered and the corresponding performance values change smoothly as shown in Figure \ref{hp_smoothness}.
Based on this observation, resulting factor matrices will exhibit smoothness property, where each row of factors is similar to its adjacent rows.
To enforce this smoothness property, we use a kernel smoothing regularization \cite{ahn2021time} applied to all factor matrices.
Given an $N$-order tensor $\T{X} \in \mathbb{R}^{I_1 \times \cdots \times I_N}$ with observed entries $\Omega$,
and a window size $S$,
we find factor matrices {$\mat{A}^{(n)} \in \mathbb R^{I_{n}\times K}$} {$,1 \leq n \leq N $} that minimizes
\begin{align} \label{eq:method:loss}
        L  = \sum_{\alpha \in \Omega}{\left({x}_{\alpha}-\sum_{k=1}^{K} \prod_{n=1}^{N}a^{(n)}_{i_{n}k}\right)^{2}}
          + \lambda \sum_{n=1}^{N}\sum_{i_{n}=1}^{I_{N}}\lVert{\vect{a}_{i_{n}}^{(n)}}-{{\tilde{\vect{a}}_{i_{n}}^{(n)}}}\rVert_{\text{2}}^{2},
\end{align}
where
\begin{equation} \label{eq:method:smooth}
{\tilde{\vect{a}}_{i_n}^{(n)}} = \sum_{i_s \in \mathscr{N}({i_n}, S)}{w(i_n,i_s)}{\vect{a}_{i_s}^{(n)}}
\end{equation}
and $\mathscr{N}(i_n, S)$ indicates adjacent indices $i_s$ of $i_n$ in a window of size $S$.
$\lambda$ is a regularization constant to adjust the effect of the smoothing.
The $\sum_{i_{n}=1}^{I_{n}}\lVert{\vect{a}_{i_{n}}^{(n)}}-{{\tilde{\vect{a}}_{i_{n}}^{(n)}}}\rVert_{\text{2}}^{2}$ term in Equation~\eqref{eq:method:loss} 
introduces the smoothness into a factor by regularizing the $i_n$th row of the factor to the smoothed vector from the neighboring rows.
The weight $w(i_n, i_s)$ denotes the weight to give to the $i_s$th row of the factor matrix.
We use the Gaussian kernel to give the weight to a row closer to the $i_t$th row should be given a higher weight.
Given a target row index ${i_n}$, an adjacent row index ${i_{s}}$, and a window size $S$,
Gaussian kernel weight is defined as follows:
		\begin{equation} \label{eq:method:kernel}
				w(i_{n},i_{s}) = \frac{\mathscr{K}(i_{n},i_{s})}{{\sum_{i_{s'} \in \mathscr{N}({i_n}, S)}} \mathscr{K}(i_{n}, i_{s'})}
		\end{equation}
		where $\mathscr{K}$ is defined by $$ \mathscr{K}(i_{n}, i_{s}) =\exp\left(-\frac{(i_{n}-i_{s})^2}{2\sigma^2}\right)$$
Note that $\sigma$ affects the degree of smoothing; a higher value of $\sigma$ imposes more smoothing.

\subsection{Proposed Ensemble Method}

Utilizing the results of multiple tensor completion methods will allow for improved accuracy and consistency. The same intuition behind ensemble machine learning models (e.g., random forest) comes into play here: Using the results of several ``weak learners'' we generate a single prediction.

\subsubsection{Simple Aggregation Functions}

First we use some simple functions to aggregate the results of several models, such as mean, median, max and min. For this method, we fully train each individual model on the sparse tensor, and each individual model makes its own prediction. After this, for each prediction, the median, for instance, is taken from all the individual's prediction. For predicting the element at index $x_i$, the ensemble model's prediction, $\text{P}_\text{Ensemble}(x_i)$, can be written as:
$\text{P}_\text{Ensemble}(x_i) = \text{median}\{\text{P}_1 (x_i), \text{P}_2 (x_i), ... \text{P}_n (x_i)\},$ 
where $\text{P}_j (x_i)$ is the jth individual model's prediction.

\subsubsection{Learned Aggregation Functions}

In addition to aggregating the individual model's results using some fixed function, we can learn a function to aggregate the results. We will experiment with training a Multi-Layer Perceptron (MLP) or Convolutional Neural Network (CNN) to aggregate the results of the individual tensor completion models. Here, each model's predictions would be the features, and each index that is being predicted would be the samples. This would be trained on the entire sparse tensor, besides the validation set. The individual models can be further trained while the learned aggregation function trains, or their training can be stopped before.

\subsubsection{Multiple Ranks for Tensor Completion}

Introducing an ensemble tensor completion algorithm also allows us to perform different rank tensor completions on a sparse tensor. For example, we can use CPD tensor completion models that use rank 5, 10, and 15 decompositions, and aggregate all their results. The intuition behind this is that we will not know the true rank of the sparse tensor we are dealing with in the real world. For this reason, aggregating the results of multiple methods with different ranks, could offer us reliability regardless of the true rank of the tensor.

\subsubsection{Splitting Data}

We can also split up the sparse tensor slightly differently for each of the individual tensors in the ensemble model. For instance, if we have multiple models, we can train each individual model using only 90\% of the entire sparse tensor. If we make sure each of the 10\% unused is different for each model, each model would have slightly different data to train on. The intuition here is for the tensor completion models to extract different information about the data, by using different pieces of the sparse tensor.

\subsubsection{Notation}

We will refer to an ensemble instance as "$\text{TenSemble-{model}}_{\_ {aggregation}}$." For example, an ensemble of CPD models, with median aggregation, will be referred to as $\text{TenSemble-CPD}_{\_ median}$.

\section{Experimental Evaluation}
\label{sec:experiments}

We use the full tensors that we generated as our ground truth. In a real world scenario, we obviously do not have access to the missing values, so there is no way to tell how well we are inferring the values. Since we generated the full tensors in this work, we will be experimenting with various types of sparse tensors (e.g. different downstream tasks, different levels of sparsity) in order to examine how well we are inferring the missing values.

\subsection{Benchmarking Tensor Completion Methods}

\begin{figure*}[!ht]
\centering
\caption{Sparse Tensor Completion MAE using 5\% observed values. Each cell represents average MAE over 5 iterations.}
\label{big_table}
\setlength{\tabcolsep}{5pt}
\begin{tabular}{lllllllllllll}
\toprule
& & NDL & & & NAS & & & QC & & & QDC & \\
\midrule
 & 1 & 2 & 3 & 1 & 2 & 3 & 1 & 2 & 3 & 1 & 2 & 3 \\
\midrule
Naive & 0.224 & 0.164 & 0.229 & 0.116 & 0.352 & 0.150 & 0.198 & 0.352 & 0.248 & 0.288 & 0.225 & 0.174 \\
CPD & 0.069 & 0.116 & \textbf{0.018} & 0.069 & 0.157 & 0.105 & 0.103 & 0.178 & 0.171 & 0.115 & 0.207 & 0.176 \\
CPD-S & \textbf{0.048} & 0.067 & 0.019 & 0.064 & 0.126 & 0.095 & 0.127 & 0.131 & 0.145 & 0.121 & 0.160 & 0.130 \\
TuckER & 0.190 & 0.085 & 0.151 & 0.126 & 0.316 & 0.120 & 0.288 & 0.262 & 0.258 & 0.276 & 0.315 & 0.269 \\
Tensor Train & 0.091 & 0.117 & 0.036 & 0.066 & 0.152 & 0.113 & 0.094 & 0.162 & 0.173 & 0.123 & 0.221 & 0.176 \\
NeAT & 0.159 & 0.119 & 0.155 & 0.108 & 0.245 & 0.120 & 0.150 & 0.222 & 0.198 & 0.185 & 0.179 & 0.146 \\
CoSTCo & 0.117 & \textbf{0.061} & 0.082 & \textbf{0.062} & \textbf{0.099} & \textbf{0.090} & \textbf{0.080} & \textbf{0.077} & \textbf{0.100} & \textbf{0.086} & \textbf{0.110} & \textbf{0.099} \\
\bottomrule
\end{tabular}

\begin{minipage}{\linewidth}
\raggedright
\footnotesize
\text{Tensors used in table:}

\text{NDL Tensors: (1) DT$\_$Dermatology$\_$828, (2) KNN$\_$Alzheimers$\_$902, (3) RF$\_$Spambase$\_$829}

\text{DL Tensors: (1) FCNN$\_$Particle$\_$ID$\_$903$\_$02, (2) FCNN$\_$Dermatology$\_$829, (3) FCNN$\_$car$\_$evaluation$\_$903}

\text{QC Tensors: (1) AND$\_$AND$\_$801, (2) AND$\_$OR$\_$801, (3) OR$\_$OR$\_$801}

\text{QDC Tensors: (1) AND$\_$AND$\_$distinct$\_$817, (2) OR$\_$AND$\_$distinct$\_$817, (3) OR$\_$OR$\_$distinct$\_$817}

\end{minipage}

\end{figure*}

As a first step, we test and compare a number of existing tensor completion methods, in order to confirm whether our goal is feasible to begin with. To do this, we fix 5\% observed values in each of our tensors (randomly sampled), and compute the error when recovering the 95\% of missing values. We can observe in Figure \ref{big_table} the performance of sparse tensor completion for our applications using a variety of tensor completion algorithms and a wide range of tasks and datasets. We compare these algorithms with a Naive method, which is just randomly sampling from the given sparse tensor (i.e. filling in the 95\% of missing values by randomly sampling the 5\% of observed values).

\subsection{Ensemble Tensor Completion Performance}
\vspace{-0.1in}
We want to explore the effect of aggregating the results of multiple tensor completion models together, in comparison with the individual tensor completion models. In these experiments we compare several tensor completion models with an ensemble, which consists of multiple tensor completion models of the same type. For example, Table \ref{CPD_vs_ensembles} displays 3 individual CPD models, of rank 1, 3, and 5, compared with different ways to aggregate these 3 models into an ensemble. We compare the individual models with their various ensembles on our four tasks: Non-Deep Learning (NDL), Neural Architecture Search (NAS), Query Cardinality (QC), and Query Distinct Cardinalty (QDC).

\begin{table}[!ht]
\centering
\caption{CPD Individual vs  Ensembles (5\% observed values)}
\label{CPD_vs_ensembles}
\setlength{\tabcolsep}{3pt}
\begin{tabular}{lllll}
\toprule
 & NDL & NAS & QC & QDC \\
\midrule
Rank 1 & .062 $\pm$ .00 & .154 $\pm$ .01 & .190 $\pm$ .06 & .149 $\pm$ .05 \\
Rank 3 & .071 $\pm$ .00 & .180 $\pm$ .02 & .207 $\pm$ .03 & .187 $\pm$ .02 \\
Rank 5 & .071 $\pm$ .01 & .158 $\pm$ .01 & .214 $\pm$ .03 & .180 $\pm$ .02 \\
\midrule
TenSemble-CPD\_$_*$ & & & & \\
\midrule
* = Median & .060 $\pm$ .00 & .151 $\pm$ .01 & .170 $\pm$ .03 & .149 $\pm$ .01 \\
* = Mean & \textbf{.058 $\pm$ .00} & .150 $\pm$ .00 & \textbf{.167 $\pm$ .03} & .145 $\pm$ .01 \\
* = MLP & \textbf{.058 $\pm$ .01} & \textbf{.127 $\pm$ .02} & .170 $\pm$ .03 & \textbf{.141 $\pm$ .02} \\
\bottomrule
\end{tabular}

\end{table}

\begin{table}[!ht]
\centering
\caption{CPD-S Individual vs Ensembles (5\% observed values)}
\setlength{\tabcolsep}{3pt}
\begin{tabular}{lllll}
\toprule
 & NDL & NAS & QC & QDC \\
\midrule
Rank 1 & .061 $\pm$ .00 & .190 $\pm$ .08 & .184 $\pm$ .10 & \textbf{.123 $\pm$ .02} \\
Rank 3 & .048 $\pm$ .00 & .177 $\pm$ .04 & .164 $\pm$ .01 & .176 $\pm$ .03 \\
Rank 5 & .047 $\pm$ .00 & .144 $\pm$ .01 & .151 $\pm$ .01 & .180 $\pm$ .02 \\
\midrule
TenSemble-CPD-S\_$_*$ & & & & \\
\midrule
* = Median & .045 $\pm$ .00 & .155 $\pm$ .03 & .141 $\pm$ .01 & .152 $\pm$ .02 \\
* = Mean & .045 $\pm$ .00 & .154 $\pm$ .03 & .148 $\pm$ .02 & .148 $\pm$ .02 \\
* = MLP & \textbf{.043 $\pm$ .00} & \textbf{.117 $\pm$ .01} & \textbf{.134 $\pm$ .00} & .156 $\pm$ .03 \\
\bottomrule
\end{tabular}

\end{table}

\begin{table}[!ht]
\centering
\caption{CoSTCo Individual vs Ensembles (5\% observed values)}
\setlength{\tabcolsep}{3pt}
\begin{tabular}{lllll}
\toprule
 & NDL & NAS & QC & QDC \\
\midrule
Rank 10 & .078 $\pm$ .01 & .124 $\pm$ .01 & .117 $\pm$ .02 & .109 $\pm$ .02 \\
Rank 20 & .067 $\pm$ .00 & .116 $\pm$ .01 & .120 $\pm$ .03 & .094 $\pm$ .01 \\
Rank 32 & .087 $\pm$ .02 & .116 $\pm$ .00 & \textbf{.099 $\pm$ .02} & .093 $\pm$ .02 \\
\midrule
TenSemble-CoSTCo\_$_*$ \\
\midrule
* = Median & .070 $\pm$ .01 & .113 $\pm$ .01 & .106 $\pm$ .02 & .089 $\pm$ .01 \\
* = Mean & .069 $\pm$ .01 & .114 $\pm$ .01 & .107 $\pm$ .02 & .089 $\pm$ .01 \\
* = MLP & \textbf{.059 $\pm$ .00} & \textbf{.105 $\pm$ .01} & .107 $\pm$ .02 & \textbf{.087 $\pm$ .01} \\
\bottomrule
\end{tabular}

\vspace{0.1cm}

\textit{Each cell is average $\pm$ standard deviation of MAE over 5 iterations.}

\end{table}

Across these three models, CPD, CPD-S, and CoSTCo \cite{liu2019costco}, we can see that aggregating the tensor completion methods in an ensemble usually improves upon the performance of the individual models. Of the aggregation functions, the MLP aggregation function seems it usually outperforms the others (mean \& median), but it is unclear how consistent this result is.

\subsection{Computational Efficiency}
\vspace{-0.1in}
Even though our proposed CPD-S and ensemble methods match but not necessarily outperform CoSTCo, CPD-S and CPD-based ensembles are more lightweight in terms of parameters to be learned and can potentially be more efficient as a result. Tables \ref{tab:timing_1} and \ref{tab:timing_2} show runtimes for the top-performing models. For the ensemble model we have divided the runtime of serially executed base models over the number of base models to simulate the ideal parallel case. Indicatively, CPD-S is either on par or faster than CoSTCo while requiring much fewer parameters and TenSemble-CPD-S was $1.9\times - 78\times$ faster than CoSTCo \footnote{For 10\% observations and higher CoSTCo often fails to converge necessitating a restart, which results in significantly higher runtime} assuming parallel execution, since each base model in the ensemble works on a sparser tensor than an individual model. As we are primarily focused on feasibility in this work, these results are highly encouraging and we defer further scalability investigation to future work.

\begin{table}[!ht]
\setlength{\tabcolsep}{4pt}
\begin{tabular}{llll}
\toprule
\% observed & CPD-S & CoSTCo & TenSemble-CPD-S\_$_{MLP}$ \\
\midrule
1\% & 0.635 $\pm$ 0.29 & 0.447 $\pm$ 0.07 & 0.737 $\pm$ 0.20 \\
2.5\% & 1.835 $\pm$ 1.48 & 1.113 $\pm$ 0.12 & 2.346 $\pm$ 1.06 \\
5\% & 2.836 $\pm$ 1.33 & 6.143 $\pm$ 3.10 & 2.612 $\pm$ 0.25 \\
10\% & 1.133 $\pm$ 0.10 & 116.460 $\pm$ 4.79 & 1.479 $\pm$ 0.11 \\
\bottomrule

\end{tabular}
\caption{Runtime in seconds for each model (training \& inference) on KNN$\_$car$\_$evaluation$\_$828. Average $\pm$ STD over 5 iterations.\label{tab:timing_1}}
\end{table}

\begin{table}[!ht]
\setlength{\tabcolsep}{3.8pt}
\begin{tabular}{llll}
\toprule
\% observed & CPD-S & CoSTCo & TenSemble-CPD-S\_$_{MLP}$ \\
\midrule
1\% & 0.173 $\pm$ 0.06 & 0.193 $\pm$ 0.04 & 0.289 $\pm$ 0.04 \\
2.5\% & 0.646 $\pm$ 0.54 & 0.366 $\pm$ 0.06 & 0.538 $\pm$ 0.21 \\
5\% & 0.541 $\pm$ 0.29 & 1.213 $\pm$ 0.31 & 0.625 $\pm$ 0.12 \\
10\% & 0.844 $\pm$ 0.20 & 8.997 $\pm$ 2.63 & 1.387 $\pm$ 0.87 \\
\bottomrule
\end{tabular}
\caption{Runtime in seconds for each model (training \& inference) on FCNN$\_$Spambase$\_$905$\_$50. Average $\pm$ STD over 5 iterations.\label{tab:timing_2}}
\end{table}

\subsection{Data Efficiency Analysis}
\vspace{-0.1in}
The rationale behind using completion to recover the entire space is to avoid evaluating a combinatorial number of potentially very expensive experiments. We are investigating in identifying the minimum number of observed entries (i.e., exact evaluations of parameter combinations) necessary in order to perform high-quality completion. In other words, we want to know how sparse our tensors can be.

\begin{figure*}[!ht]

    \caption{ \label{sparsity_bars} Error vs. levels of sparsity for various models across all four tasks. CoSTCo \cite{liu2019costco} \& its ensemble consistently perform the best with little observed entries. Closer to 5\% observed entries, the rest of the models seem to catch up. Our proposed CPD-S \& $\text{TenSemble-CPD-S}_{\_ MLP}$, however, has similar performance to CoSTCo \cite{liu2019costco}, without using as many parameters as a CNN.}
    
    \centering
    \subfigure[Non-Deep Learning Tensor]{
        \includegraphics[width=0.48\linewidth]{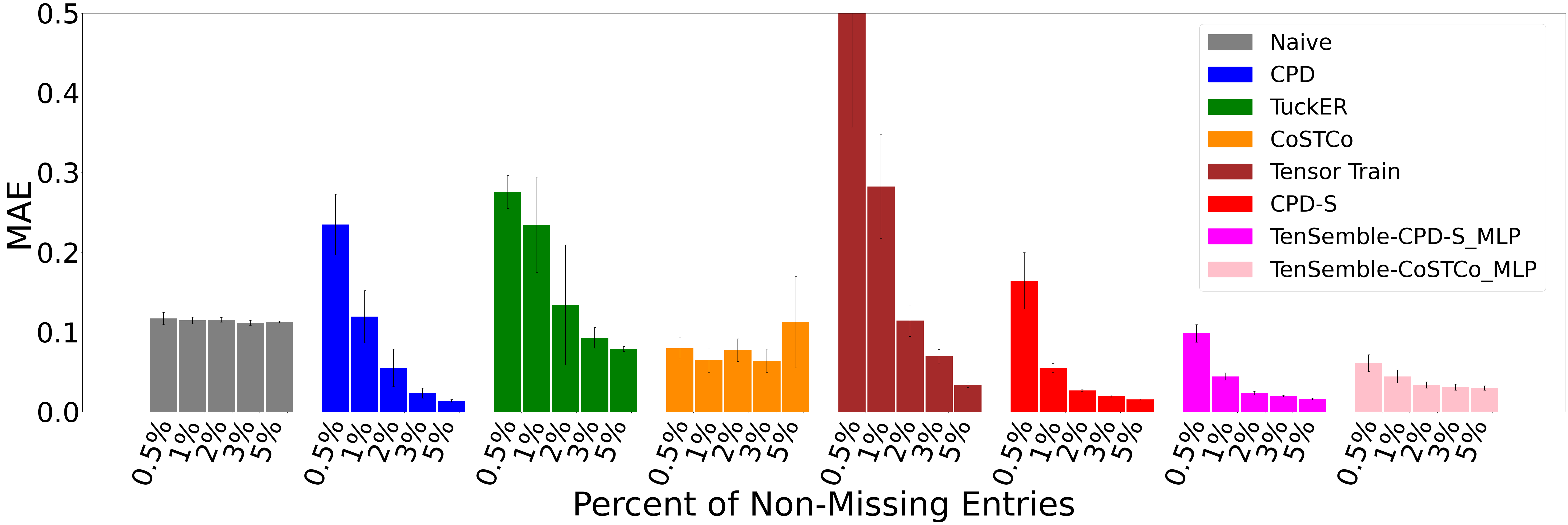}
    }
    \subfigure[Neural Architecture Search Tensor]{
        \includegraphics[width=0.48\linewidth]{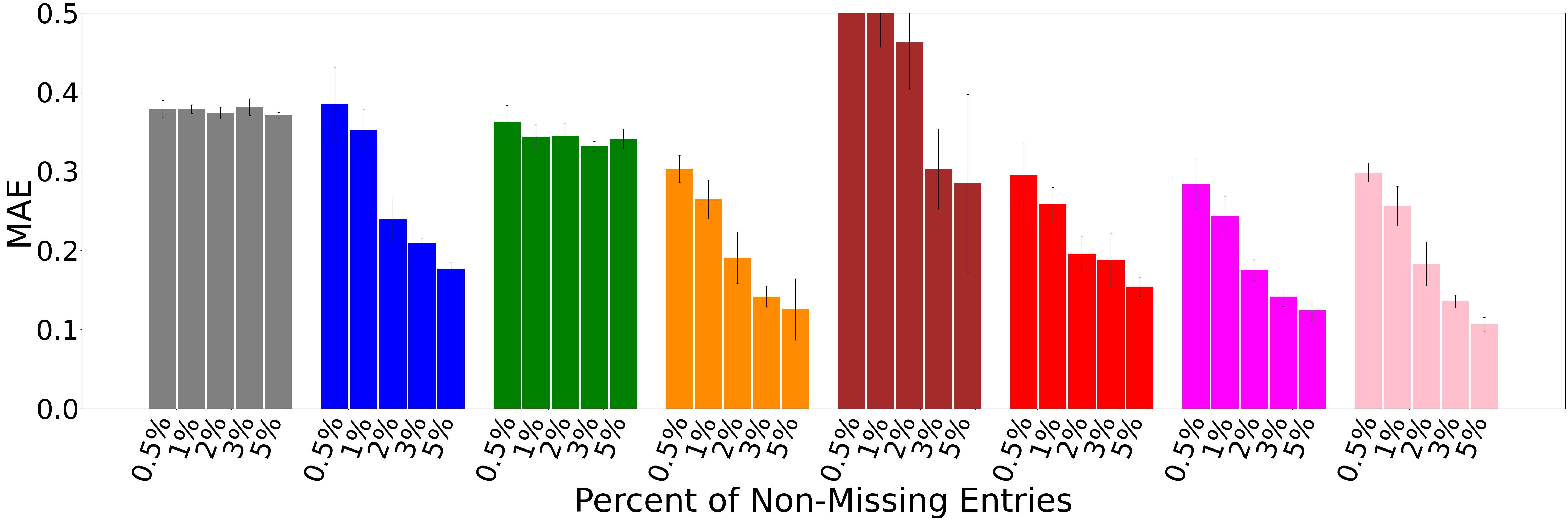} 
    } 
    \subfigure[Query Cardinality Tensor]{
        \includegraphics[width=0.48\linewidth]{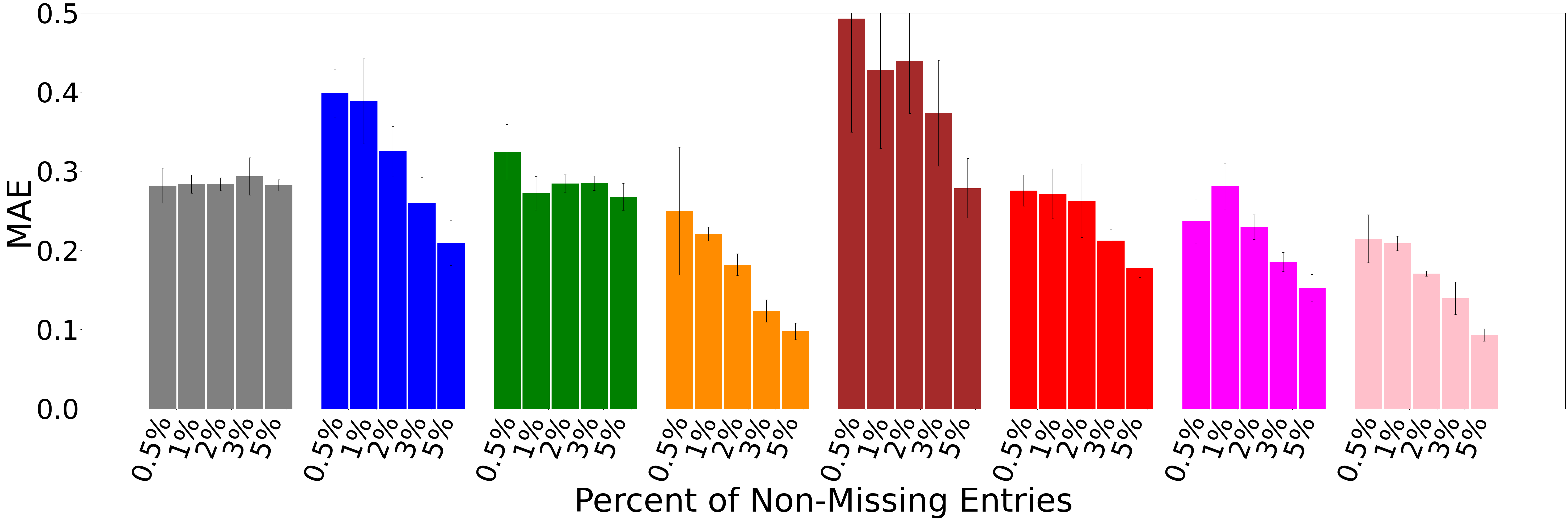}
    }
    \subfigure[Query Distinct Cardinality Tensor]{
        \includegraphics[width=0.48\linewidth]{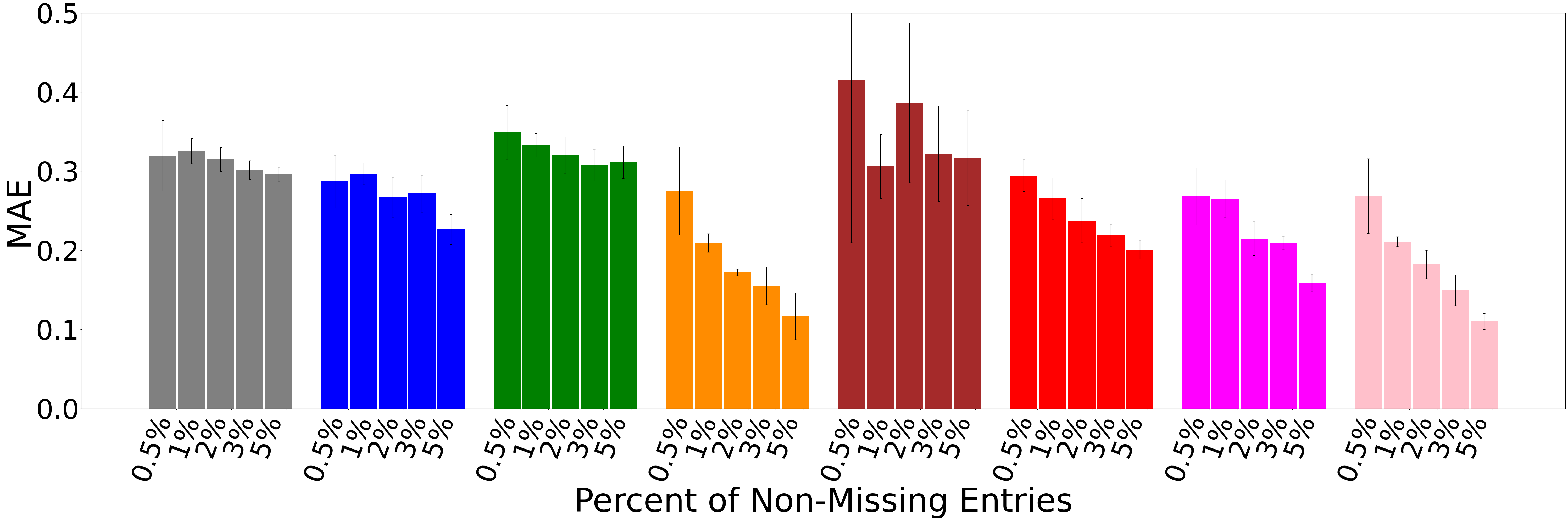} 
    } 

\end{figure*}

Figure \ref{sparsity_bars} displays the Mean Absolute Error (MAE) of tensor completion on the Y-axis, and the percent of observed entries in the sparse tensor on the X-axis (grouped by tensor completion model). The graphs show the performance of various tensor completion methods with respect to the sparsity of the tensor. One distinct pattern to be seen is the difficulty of sparse tensor completion with relation to the rank of the tensor. Query Cardinality and Non-Deep Learning tensors seem the easiest to complete, due to the low rank of the tensors. The Neural Architecture Search tensors seem harder to complete due to their high rank.

For many tasks, CoSTCo \cite{liu2019costco} has the least error with more missing values; however, our CPD-S Ensemble model gives very similar performance on most of the tensors. Interestingly though, this is not the case for the Query Cardinality Tensors.
Across a variety of Non-Deep Learning and Neural Architecture Search tensors, though, we see that the CPD-S Ensemble gives very similar performance of CoSTCo, without needing the Convolutional Neural Network .

\subsection{CPD-S Smoothness Sensitivity}
\vspace{-0.1in}
When adding a smoothness constraint to our CPD tensor completion, we can choose to enforce this constraint more strictly or more loosely. This is done by increasing or decreasing the smoothness constraint's coefficient term, lambda, in our loss function. Note that when lambda = 0, the loss function is just the base loss function (MSE).

\begin{figure}[!htp]
    \centering
    \caption{CPD-S Error with different lambda coefficient values, compared with regular CPD \& Naive methods. These graphs display that a positive lambda value (enforcing smoothness constraint) almost always decreases the error in the scope of our application.}
    \includegraphics[width=0.85\linewidth]{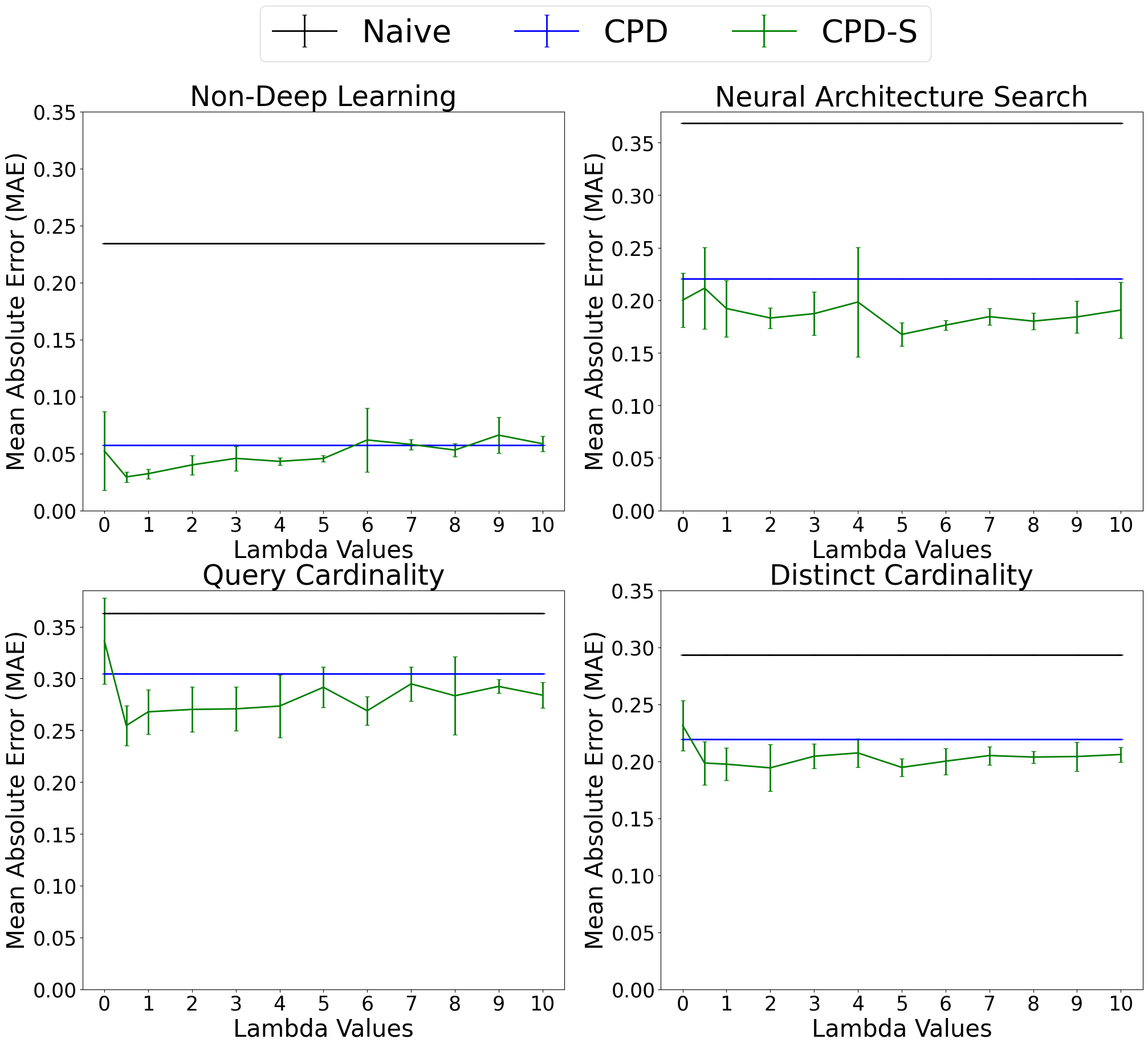}
    \label{fig:cpd_s_sensitive}
    
\end{figure}

We observe in Figure \ref{fig:cpd_s_sensitive} that a positive lambda value almost always helps with tensor completion in our application. In some cases, a smaller lambda value is more effective, but this does not seem to be true for all cases.

\subsection{Investigating the Latent Structure of the Data}
\vspace{-0.1in}
In our tensor completion algorithms, we assume there is structure in the tensors that we are completion, otherwise we could not complete them. However, it is unclear whether or not we can assume a strictly low-rank structure. In the real world, we will not be able to find out what the exact rank of our sparse tensor is, because to the all of the missing values. It is for this reason that it is important to conduct some analysis with respect to the rank of the tensors we are completing, since we do have the completed dense tensors.

\begin{figure}[!htp]
    \centering
    \caption{These graphs display the normalized (0, 1) error when decomposing and reconstructing a dense tensor of each task, with respect to the decomposition rank.}
    \includegraphics[width=0.85\linewidth]{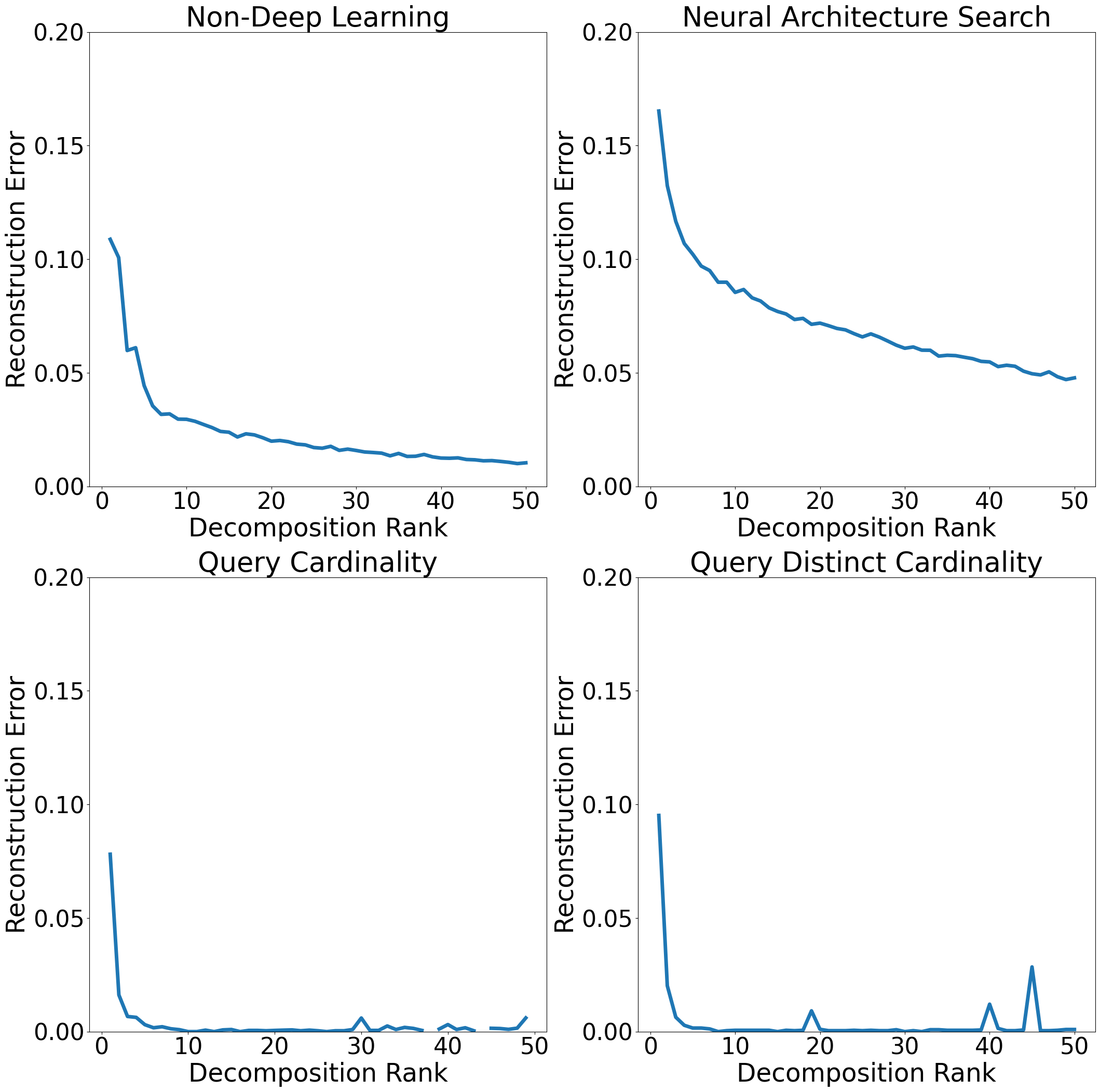}
    \label{fig:tensor_ranks}
    
\end{figure}

In Figure \ref{fig:tensor_ranks}, we perform TensorLy's PARAFAC decomposition \cite{kossaifi2019tensorly} on our dense tensors. From this decomposition, we reconstruct the entire dense tensor back again, and record this error. When doing this decomposition and reconstruction across many different ranks, we observe what ranks are needed to accurately reconstruct the tensor. This gives us an estimation on the tensor's rank. For example, the pictured Query Cardinality tensor only needs a low-rank decomposition to accurately estimate the values of the entire dense tensor. On the other hand, the Neural Architecture Search tensor requires a much higher rank decomposition to be able to estimate the values accurately. 

Generally we observe this trend where Query tensors tend to be very low rank, Non-Deep Learning tensors are slightly higher rank, and Neural Architecture Search tensors are significantly higher rank. This observation comes in contrast to previous work \cite{deng2022new, yang2020automl, rebelloefficient}, that makes a strong low-rank assumption, which appears to not always be the case in the diverse set of scenarios we explore here, and futher justifies the good performance of methods like CoSTCo which do not make strong low-rank assumptions.

\begin{figure}[!htp]
    \centering
    \caption{Completion error for low and high rank tensors. The top right is for a query cardinality tensor, top left graph for a SVM tensor, and the bottom two for Neural Architecture Search tensors.}
    \includegraphics[width=0.85\linewidth]{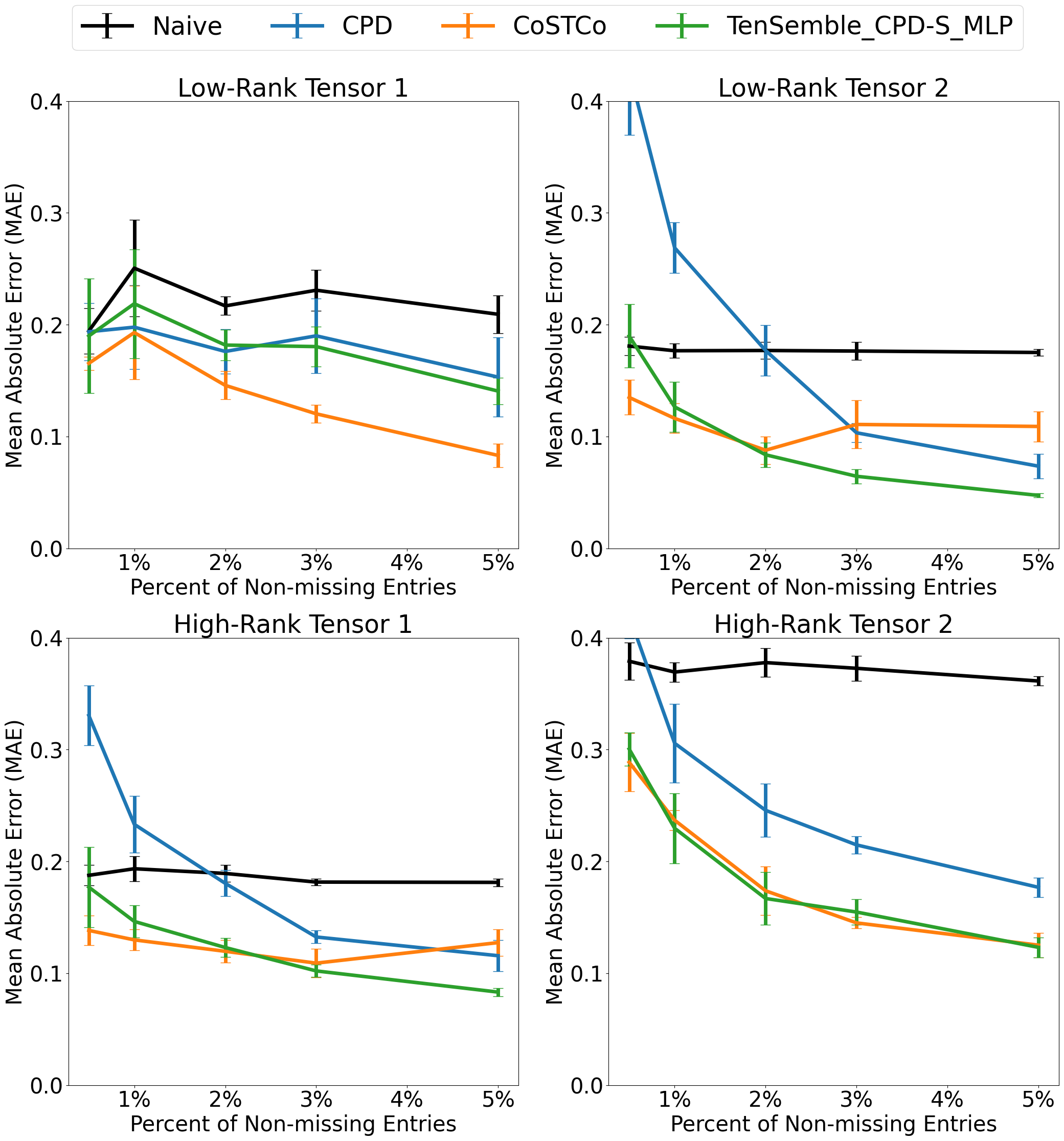}
    \label{fig:low_high_rank_completion}
    
\end{figure}

Previous work on Tensor Completion for AutoML assumes a low-rank structure of the tensor \cite{deng2022new, yang2020automl, rebelloefficient}; however, we experimentally verify the viability of tensor completion even for higher rank tensors. Not only does this allow us to extend our work into Neural Architecture Search (which tends to produce higher rank tensors), but it also gives us robustness when we do not know the exact rank of our sparse tensor. Figure \ref{fig:low_high_rank_completion}  shows results in low and high rank tensors.

Finally, we would like to note that low-rankness of a given tensor, as measured here, does not necessarily imply that the task is trivial. This is because our rank assessment is done using the full tensors, while in reality we would have to estimate that low-rank space from a small amount of observations, which is challenging.

\subsection{Cross-dataset Completion}
\vspace{-0.1in}
\begin{figure}[!htp]
    \centering
    \caption{Introducing a dataset mode, for non-deep learning and neural architecture search tensors. The dataset mode is to simulate where we might already have the results of previous tasks. Note the other slices corresponding to the other datasets have 15\% observed entries for this experiment.}
    \includegraphics[width=0.85\linewidth]{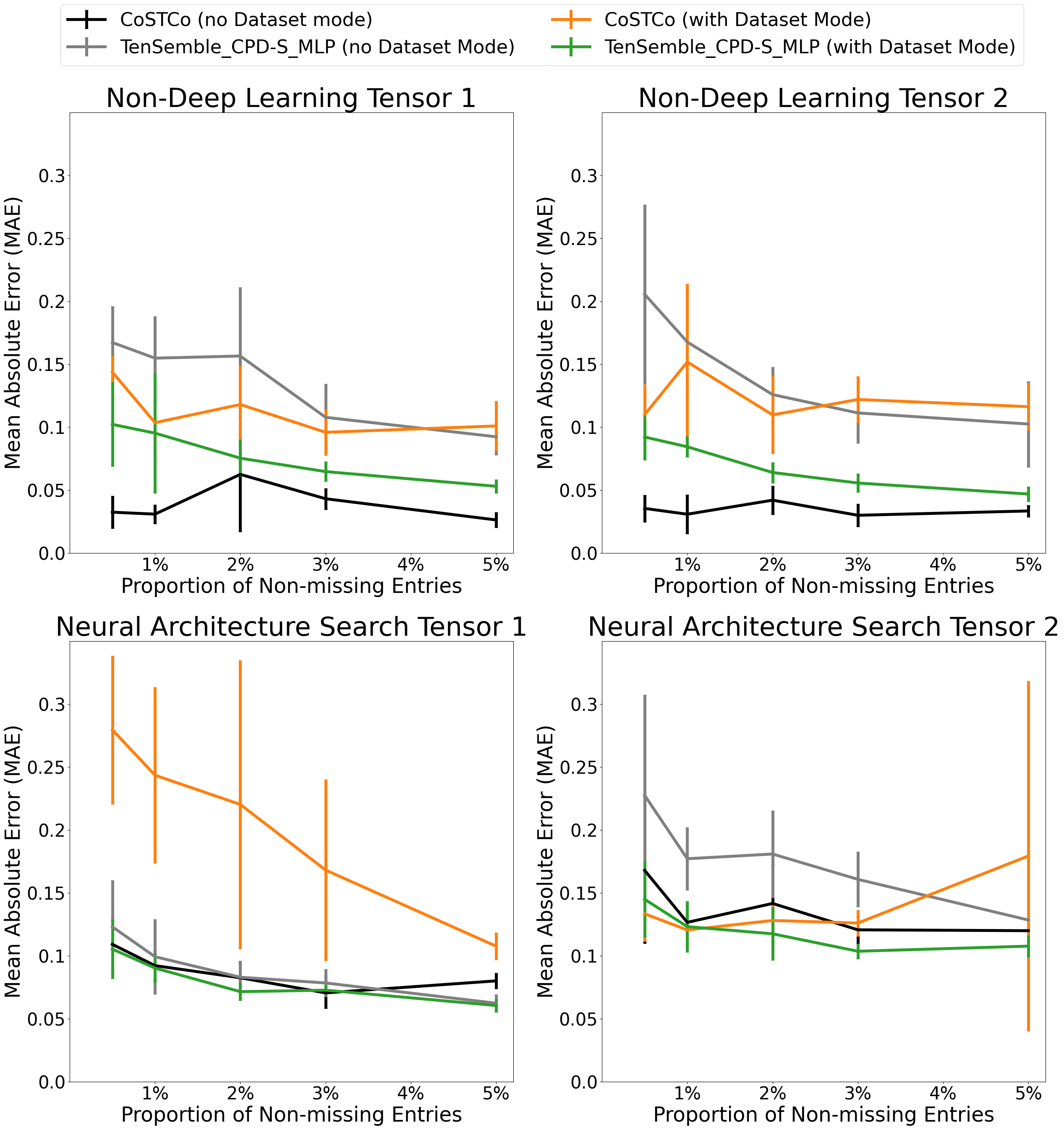}
    \label{fig:dataset_mode_graph}
    
\end{figure}

Taking advantage of the flexibility and extendibility of tensor modeling, we ask the question: can we introduce a ``dataset'' mode and conduct joint completion across different downstream datasets? Essentially, can we transfer results from one dataset to another using our formulation?
Figure \ref{fig:dataset_mode_graph} shows the effects of introducing a dataset mode. Interestingly enough, CoSTCo \cite{liu2019costco} actually seems to suffer from introducing the dataset mode, likely due to overfitting to the other datasets' tensors. On the other hand, we observe that our TenSemble-CPD-S\_$_{MLP}$ model can benefit significantly from the introduction of a dataset mode.
\section{\textbf{Related Work}: Tensor Methods for AutoML}
\label{sec:related}
\vspace{-0.1in}

There exists immediately related work that addresses the exact issue of using Sparse Tensor Completion for AutoML \cite{deng2022new, yang2020automl, rebelloefficient}. In contrast to those works, we relax the strict low-rank assumptions. This is to our benefit as we do come across a variety of Neural Architecture Search tensors that are quite high rank. This also adds to the robustness of this application since it may be difficult to tell the exact rank of the tensor when we only have a small fraction of observed entries. Finally, we go beyond hyperparameter optimization, as we take a broader approach and use tensor completion methods for a variety of data science tasks. 

\section{Conclusions}
\vspace{-0.1in}
We make the following high-level contributions:
\begin{itemize}
    \item We cast a number of diverse data science tasks, such as hyperparameter optimization, neural architecture search, and query cardinality estimation, which currently act as computationally intensive bottlenecks in building data science pipelines under the unifying umbrella of tensor completion.
    \item We extensively compare state-of-the-art tensor completion methods and confirm feasibility of our proposition.
    \item Inspired by the underlying structure that emerges in the data  and by the observed behavior of existing completion methods, we propose a framework of novel tensor completion methods that are able to achieve state-of-the-art performance in our tasks.
    \item We publicly release our code and benchmark datasets, empowering further research in this direction.
\end{itemize}

\section*{Acknowledgements}
\vspace{-0.1in}
\small{We would like to thank Kuntal Pal for initial discussions. Research was supported by the National Science Foundation under REU Site grant no. OAC 2244480, CAREER grant no. IIS 2046086 and CREST Center for Multidisciplinary Research Excellence in Cyber-Physical Infrastructure Systems (MECIS) grant no. 2112650.}%

\balance
\bibliographystyle{IEEEtran}
\bibliography{bib/vagelis_refs}

\end{document}